\pdfoutput=1%For ARXIV
\documentclass[letterpaper]{article} % DO NOT CHANGE THIS
\usepackage{aaai24}  % DO NOT CHANGE THIS
\usepackage{times}  % DO NOT CHANGE THIS
\usepackage{helvet}  % DO NOT CHANGE THIS
\usepackage{courier}  % DO NOT CHANGE THIS
\usepackage[hyphens]{url}  % DO NOT CHANGE THIS
\usepackage{graphicx} % DO NOT CHANGE THIS
\usepackage{natbib}  % DO NOT CHANGE THIS AND DO NOT ADD ANY OPTIONS TO IT
\usepackage{caption} % DO NOT CHANGE THIS AND DO NOT ADD ANY OPTIONS TO IT
\usepackage{algorithm}
\usepackage{algorithmic}
\usepackage{amsmath}
\usepackage{multirow}

\usepackage{newfloat}
\usepackage{listings}
\DeclareCaptionStyle{ruled}{labelfont=normalfont,labelsep=colon,strut=off} % DO NOT CHANGE THIS
\floatstyle{ruled}
\newfloat{listing}{tb}{lst}{}
\floatname{listing}{Listing}
\title{{DLCA-Recon: Dynamic Loose Clothing Avatar Reconstruction from 

Monocular Videos}}
\author {
    Chunjie Luo,
    Fei Luo$^*$,
    Yuseng Wang,
    Enxu Zhao,
    Chunxia Xiao\thanks{Fei Luo and Chunxia Xiao are co-corresponding authors.}
}
\affiliations {
    \textsuperscript{\rm 1}School of Computer Science, Wuhan University, Wuhan, China\\
    luochunjie@whu.edu.cn, luofei@whu.edu.cn, wangyusen@whu.edu.cn, zhaoenxu@whu.edu.cn, cxxiao@whu.edu.cn
}
\usepackage{bibentry}
\begin{document}

\maketitle

\begin{abstract}
Reconstructing a dynamic human with loose clothing is an important but difficult task. To address this challenge, we propose a method named DLCA-Recon to create human avatars from monocular videos. 
The distance from loose clothing to the underlying body rapidly changes in every frame when the human freely moves and acts. Previous methods lack effective geometric initialization and constraints for guiding the optimization of deformation to explain this dramatic change, resulting in the discontinuous and incomplete reconstruction surface.
To model the deformation more accurately, we propose to initialize an estimated 3D clothed human in the canonical space, as it is easier for deformation fields to learn from the clothed human than from SMPL.
With both representations of explicit mesh and implicit SDF, we utilize the physical connection information between consecutive frames and propose a dynamic deformation field (DDF) to optimize deformation fields. DDF accounts for contributive forces on loose clothing to enhance the interpretability of deformations and effectively capture the free movement of loose clothing. Moreover, we propagate SMPL skinning weights to each individual and refine pose and skinning weights during the optimization to improve skinning transformation. Based on more reasonable initialization and DDF, we can simulate real-world physics more accurately. Extensive experiments on public and our own datasets validate that our method can produce superior results for humans with loose clothing compared to the SOTA methods.
\end{abstract}

\section{Introduction}

Reconstructing full-body 3D human models is an important research topic in computer graphics. It has many applications in AR/VR \cite{bao2022deep,cao2022dgecn,cao2023dgecn++}, virtual try-on, and video game industry. Traditionally, high-fidelity human reconstruction requires multi-camera systems, controlled studios, and long-term works of talented artists, making it expensive and highly specialized. Along with the emergence of new applications like digit human in the Metaverse, it demands lightweight and convenient reconstruction solutions to create 3D digital avatars for complex human motions and diverse manners of dressing. %They also need to be applicable in both indoor and outdoor scenarios. 

%It is one of the dressing fashions for people to wear loose-fitting clothes. 
Compared to close-fitting wear, it is more difficult to reconstruct dynamic humans with loose clothing, due to the high freedom of body motions, the appearance diversity, and the deformation randomness of loose clothes. Traditional methods based on explicit mesh are restricted by fixed topologies and resolutions \cite{alldieck2018video,guo2021human}. Recent methods based on the implicit neural representation for monocular human reconstruction have achieved compelling results \cite{saito2019pifu,huang2020arch,he2020geo,gropp2020implicit,zheng2021pamir,xiu2022icon}. These methods can handle arbitrary topologies, enabling the representation of various clothing. However, they require high-quality 3D supervision data and NeRF-based methods usually produce noised geometry. %Therefore, reconstructing dynamic avatars from readily available in-the-wild videos is a challenging task. 
\begin{figure}[t]
\centering
\includegraphics[width=1.0\columnwidth]{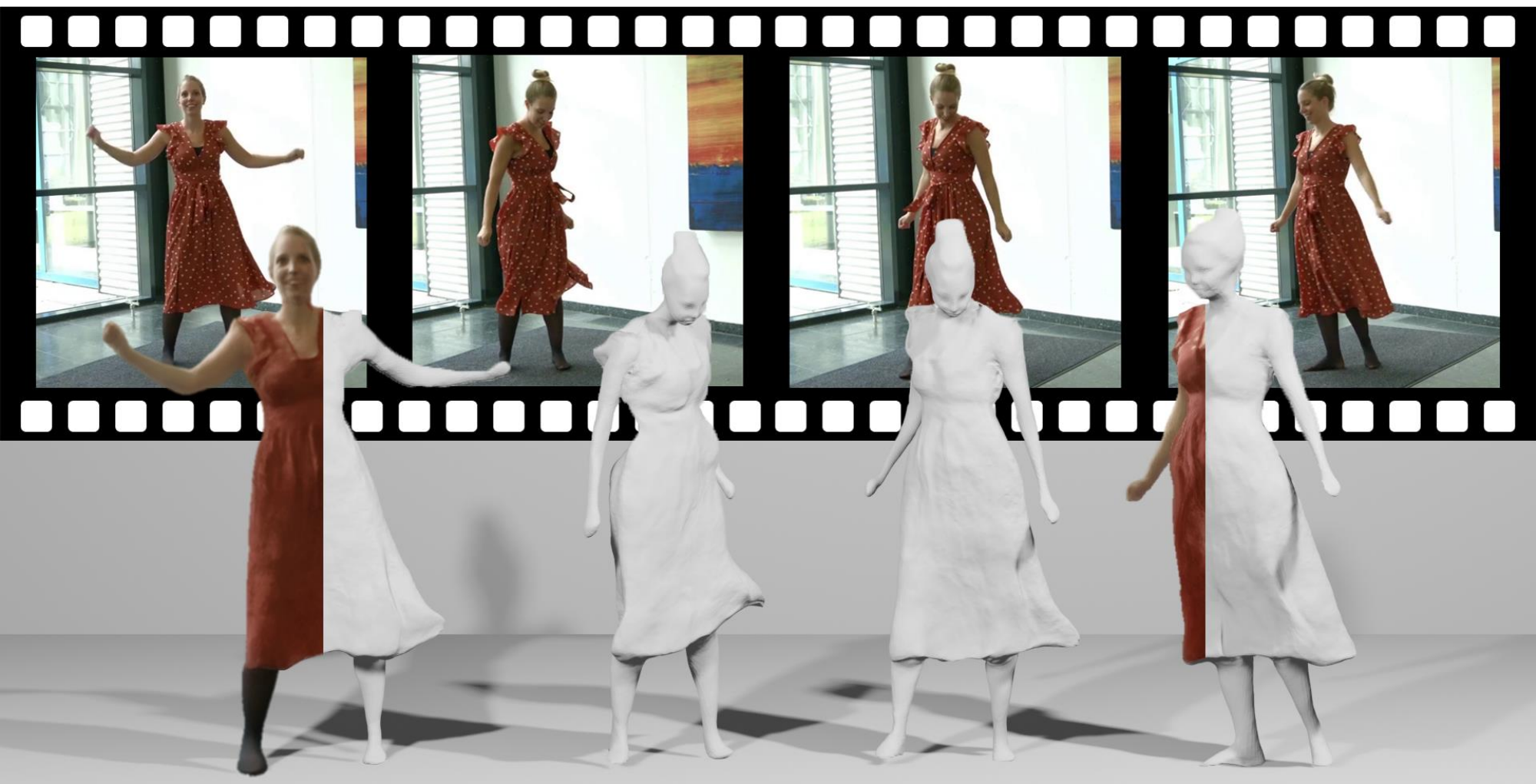} 
\caption{A dynamic loose clothing avatar created by our DLCA-Recon method from a monocular video.}
\label{introduction}
\end{figure}

In addition, there are methods working in a frame-by-frame manner, but they fail to recover invisible parts. Directly regressing 3D surfaces from images is an alternative way \cite{he2020geo,huang2020arch,he2021arch++,peng2021animatable,peng2021neural,weng2022humannerf,li2023monocular,luo2023sparse}. They struggle with out-of-distribution of poses and shapes and cannot carry out temporally continuous 3D reconstructions. SelfRecon \cite{jiang2022selfrecon} combines explicit and implicit representation to reconstruct a temporally consistent 3D clothed human from a video. But it is restricted to tight clothing and self-rotation movement.

To tackle creating a spatio-temporally coherent 3D model for a human with loose clothing and free movement, we propose a new method DLCA-Recon based on SelfRecon \cite{jiang2022selfrecon}. When dealing with the diverse topologies of loose clothing, we initialize a clothed human 3D model in the canonical space using a single image avatar creation method \cite{xiu2023econ}, unlike SMPL+D initialization of other methods. Such a different starting point decreases the gap between initialization and subsequent deformation iterations, leading to higher accuracy of mapping canonical points to the current frame space \cite{chen2021snarf,zheng2022avatar}. During the iterations of reconstruction, we focus on the dynamics in the non-rigid field and carefully update weights in the skinning transformation field to better explain and simulate clothing movement. We propose a dynamics method DDF to model the influence of related forces on the deformation of loose clothing, which enhances deformation interpretability and enables DLCA-Recon to more accurately simulate real-world physics. Moreover, we optimize human poses and manage the overall network optimization to prevent training collapse. 
Our contributions could be summarized as follows,
\begin{itemize}
\item We propose to use an estimated human geometry as mesh initialization in the canonical space, which could better guide the SMPL weight propagation to the body and clothes. We especially fine-tune body pose and skinning weights to improve skinning transformation; 
\item We propose a dynamic deformation field (DDF) to account for all major contributive forces, which could effectively model the free movement of body and clothes;
\item Extensively experimental evaluations on benchmark datasets and our captured monocular videos demonstrate that our method outperforms existing methods. We provide a more robust spatial-temporal reconstruction method for 3D dynamic avatars with loose clothing.
\end{itemize}

\section{Related Work}

\textbf{Clothed Human Reconstruction from Single-View Image.} 
Traditional human reconstruction often adopts a parametric model, $e.g.$ SMPL \cite{loper2015smpl} or SCAPE \cite{anguelov2005scape} and only recover a naked 3D body \cite{joo2018total,kanazawa2018end}. Many methods use ``SMPL+D'' to represent 3D clothed humans \cite{alldieck2018video, alldieck2019learning, alldieck2019tex2shape,zhu2019detailed, ma2020learning, xiang2020monoclothcap}. However, this ``body+offset" approach is not flexible enough to model loose clothing like dresses and skirts. 

Recent methods introduce implicit representation to increase topological flexibility. PIFu and PIFuhd \cite{saito2019pifu, saito2020pifuhd} extract pixel-aligned spatial features from images to implicit surface function. Two methods do not leverage knowledge of the human body structure, resulting in overfitting the body poses in training data. Consequently, they fail to generalize the 3D model to novel poses and produce shapes with broken or disembodied limbs.  

To address these issues, several methods \cite{huang2020arch,he2021arch++,zheng2021deepmulticap,zheng2021pamir,liao2023high} combine parametric body models with implicit representations. To further generalize to unseen poses, ICON \cite{xiu2022icon} regresses shapes from locally queried features. These approaches enhance robustness to unseen poses but still have not enough generalization ability to various, especially loose, clothing topologies.  Recently, ECON \cite{xiu2023econ} directly generates the clothed human from bilateral normal integration, enabling loose-fitting clothing reconstruction. But it tends to output bent legs and incorrect thickness of human. %Besides, the method is extremely dependent on normal. Once the normals are predicted wrong, the results tend to be wrong.

As these methods only consider single-image reconstruction, they cannot produce temporally consistent results. The results can be wrong in other views. Moreover, these methods require a large amount of 3D scanned ground truth to ensure generalization capability.

\textbf{Clothed Human Reconstruction from Monocular Video.} Traditional methods require personalized rigged templates as prior and track the pre-defined human model based on 2D observations \cite{xu2018monoperfcap,habermann2019livecap,habermann2020deepcap}. These methods require pre-scanning and manual rigging, unsuitable for lightweight applications. Some explicit methods \cite{alldieck2018video,guo2021human,casado2022pergamo,moon20223d} omit personalized rigged templates but are still limited to a fixed resolution and topologies. Some methods \cite{pons2017clothcap,tiwari2020sizer,xiang2022dressing,casado2022pergamo} reconstruct the clothing as a separate layer over the body with high-quality 3D clothing supervision. 

Some approaches introduce implicit methods to capture the details and facilitate 3D body reconstruction. NeuralBody \cite{peng2021neural} represents dynamic human NeRF based on SMPL. HumanNeRF \cite{weng2022humannerf}  extends articulated NeRF to improve novel view synthesis. NeuMan \cite{jiang2022neuman}  further adds a scene NeRF model. These methods model the geometry with a density field, yielding low-fidelity and spatial-temporal inconsistent human reconstruction. SelfRecon \cite{jiang2022selfrecon} combines explicit and implicit representation to reconstruct temporally consistent 3D clothed humans, but it could not reconstruct humans in loose clothing and free motion. Vid2avatar \cite{guo2023vid2avatar} utilizes self-supervised scene decomposition to achieve temporally consistent human reconstruction, but it is not special for loose clothing.

\begin{figure*}[t]
\centering
\includegraphics[width=1.0\textwidth]{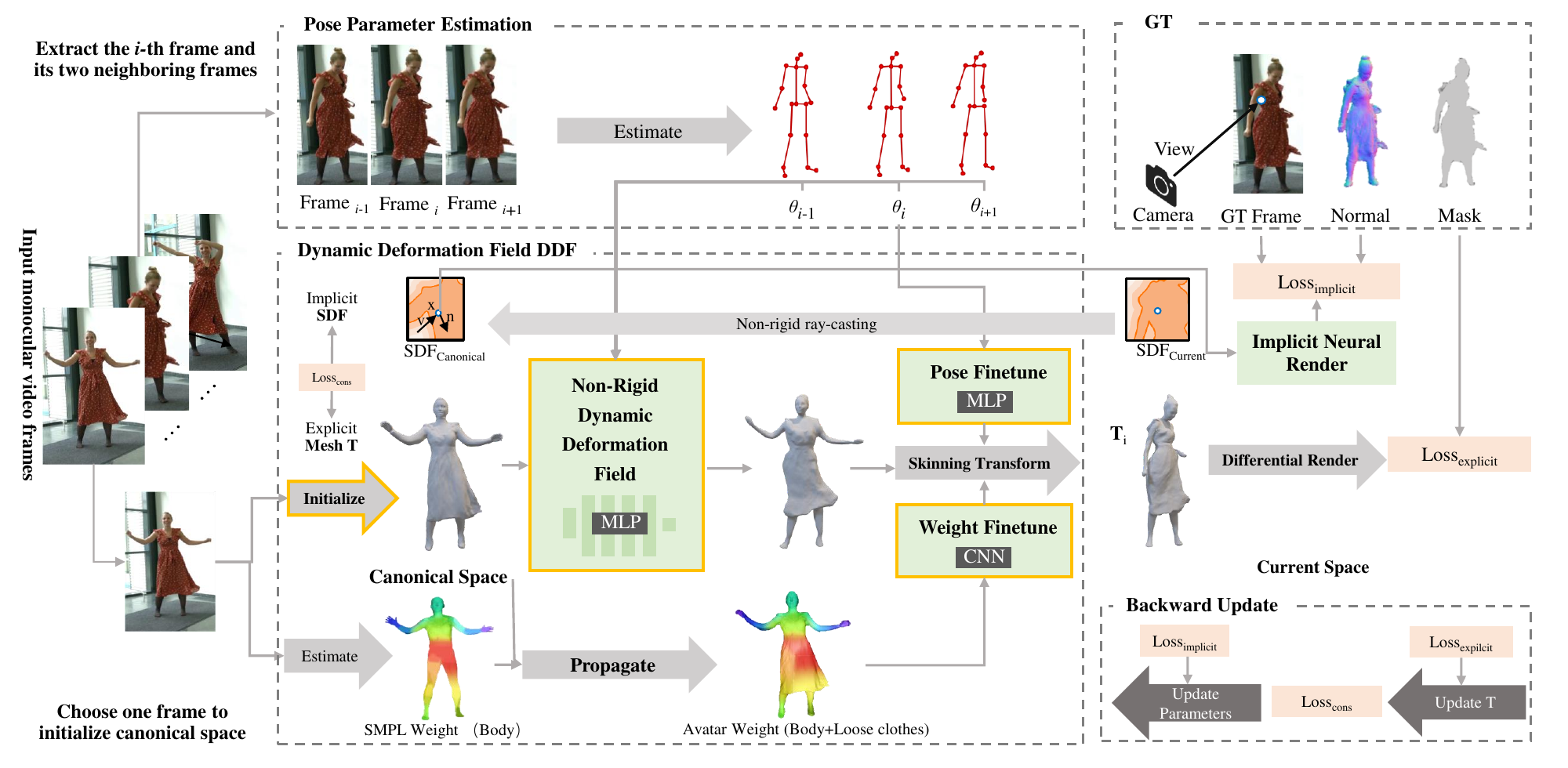} % Reduce the figure size so that it is slightly narrower than the column.
\caption{Diagram of DLCA-Recon. Inputting a monocular video, DLCA-Recon first initializes a 3D human mesh in canonical space. Through maintaining double representations of explicit mesh and implicit SDF for the avatar, DLCA-Recon handles the current \textit{i}-th frame by pose parameter estimation, dynamic deformation field DDF, and loss calculation. In the backward update procedure, the explicit representation loss updates the mesh $T$ in the canonical space, and then the consistency loss aligns the explicit representation and the implicit representation. Finally the implicit loss updates all learnable parameters. Gray arrow denotes an operation. Green rectangle represents a certain network and yellow border parts are our proposed modules.  %Our method takes a monocular video as input, and outputs a human avatar. At the  beginning, we diffuse the SMPL weight to the initial mesh. It is computed only once in the whole optimization. We input points in canonical space and poses of three adjacent frames to the non-rigid deformation field. Then, we input the deformed points and  optimized pose in current space to skinning transformation field. The weights of the skinning transformation field are continuously updated in the training time. After that, we obtain the mesh corresponding to the current frame. We minimize the color loss is calculated from the RGB obtained by implicit neural rendering. We use differentiable mask loss to recover the overall shape. To simplify the framework overview, We illustrate non-rigid deformation field in Figure \ref{non_rigid}.
}
\label{pipeline}
\end{figure*}

\section{Methodology}
%DL-Avatar aims to reconstruct high-fidelity and space-time coherent clothed human with arbitrary garments from monocular videos in the wild. 
The proposed DLCA-Recon method is schematically illustrated in Figure \ref{pipeline}. Like SelfRecon \cite{jiang2022selfrecon}, 
we jointly optimize the explicit and implicit representation. Given a monocular video, we first randomly choose one frame and define the canonical human representation in both explicit mesh and implicit signed-distance field (SDF). During training, DLCA-Recon estimates the pose parameters for each frame and its two neighboring frames, which would be inputted into the forward deformation to deform the 3D human model from canonical space to each frame’s space. It consists of the non-rigid dynamic deformation field and the optimized skinning transformation field. The non-rigid dynamic deformation field (DDF) aims to capture the movement of loose clothing and generate spatial-temporal coherent explicit meshes.  Finally, we utilize mask loss to control the shape of explicit mesh and improve details through normal loss and color loss.
 
\subsection{Non-Parametric Initialization}
%The shapes and appearances of loose clothing pose a challenge when using SMPL initialization alone. 
Employing a coarse human body initialization provides enhanced adaptability to different types of loose clothing, 
%resulting in a more accurate initial estimation. 
so we use a coarse human representation instead of SMPL in canonical space. Specifically, we select a frame from the monocular video and feed it into ECON \cite{xiu2023econ} to obtain a geometry as the initialization. Then, to represent high-fidelity geometry, we define a canonical SDF $S$ by an MLP $f$ of the geometry using IGR \cite{gropp2020implicit}: 
\begin{equation}
S=\left \{ x_c \mid f(x_c)=0  \right \} , \label{sdf}\\
\end{equation}
where $x_c$ is the vertex in canonical space.

Moreover, due to different cameras used in ECON and weak supervision on 2D images, the geometry estimated from ECON may have inconsistent scales and misaligned positions in canonical space. In 3D reconstruction, providing an initialization that is dimensionally and positionally incorrect could lead to convergence challenges, instability, and shape distortion. So it is necessary to predict an SMPL in canonical space, as well as align and scale the initialization with the SMPL.

\subsection{Dynamic Deformation Field}
The deformation of a clothed human cannot be fully represented by skinning transformation. Following prior works \cite{jiang2022selfrecon,weng2022humannerf,guo2023vid2avatar}, we decompose the deformation field into a non-rigid deformation field and a skinning deformation field. Meanwhile, based on the force analysis of clothing, we propose a dynamic deformation field (DDF). Our new deformation field consists of a non-rigid dynamic deformation field and an 
initialization-based skinning transformation field.
Given a monocular video depicting a clothed person in free motion, we generate per-frame SMPL \cite{loper2015smpl} pose parameters
$\left \{ \theta_i|i=1,...,N  \right \} $ using PyMAF \cite{zhang2021pymaf}.

\begin{figure}[htb]
\centering
\includegraphics[width=1.0\columnwidth]{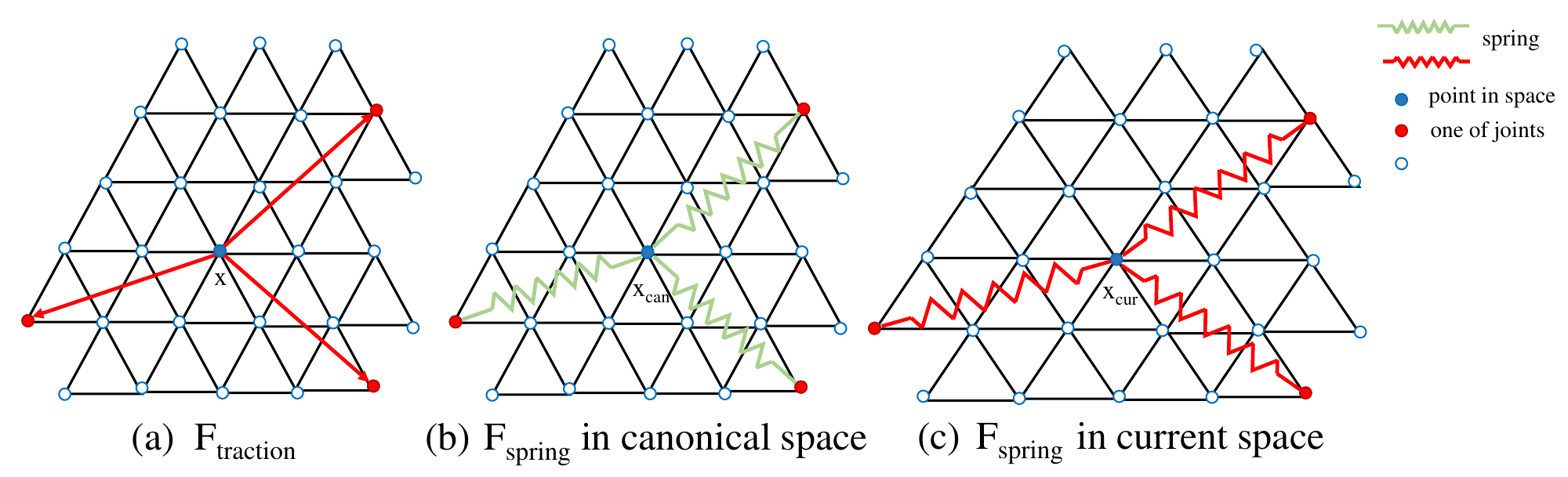} 
\caption{Forces on the clothing. We represent clothing as a combination of many triangular surfaces. Blue points on the triangles are points of the garment geometry, and red points indicate human joints. (a) shows the $F_{traction}$ that the point $x$ receives from one of the joints. This force can be directly represented as a line from point $x$ to the joint. (b) shows the state of the spring model in canonical space. We can simply understand that the green spring is in a relaxed state. In this case, the spring length sets the default to the original length. (c) shows the spring model in the current space. The red spring is stretched due to human movement.}
\label{non_rigid}
\end{figure}

\subsubsection{Non-Rigid Dynamic Deformation Field.}
According to dynamics, the motion of an object is related to the forces on it. Physics-based cloth simulation analyzes the internal and external forces acting on each vertex. We primarily address the free motion of the human in nature scenes. In this case, the forces acting on each vertex include gravity, traction, and friction resulting from human movement, as well as air resistance. In addition, internal force affects each other between the vertices. To simplify the representation, we compute the total force acting on a clothing vertex at time $t_i$ using vector operations. According to Newton's second law, $F=ma$, the traction acting on a vertex at time $t_i$ can be expressed as
\begin{eqnarray}    \label{eq}
F_{traction_i}&=&ma_i =m\frac{\Delta v_i}{\Delta t} =m\frac{\Delta s_i}{\Delta t^{2}} \nonumber    \\
~&=&m\frac{x_{i+1}-x_{i-1}}{2\Delta t^{2} } ,
\end{eqnarray}
where $a_i$ is the instantaneous acceleration and $v_i$ is the instantaneous velocity at the current frame. The mass $m$ and time step $\Delta t$ remain constant in a monocular video. Thus the force acting on a vertex at time $t_i$ is determined by $x_{i+1} –  x_{i-1}$. Since the traction force is mainly generated by human motion, $x_{i+1} –  x_{i-1}$ can be expressed as a positive correlation of $J_{i+1} –  J_{i-1}$. $J_{i+1} –  J_{i-1}$ is the distance between the human joints of the subsequent video frames. Therefore, the traction can be presented as follows:
\begin{equation}
F_{traction_i}\propto (J_{i+1} –  J_{i-1}). \label{XX}\\
\end{equation}
Once the pose $\theta$ and translation $T$ are available, the joints $J$ can be known. So the traction is:
\begin{equation}
F_{traction_i}\propto (\theta_{i+1},T_{i+1}, \theta_{i-1},T_{i-1}). \label{XX}\\
\end{equation}

The internal forces of the clothing primarily utilize a spring model, following Hooke's law $F_{spring} = -k \Delta x $. To calculate the internal force, we need to get $ \Delta x $, which represents the spring's length change. As Figure \ref{non_rigid} shows, in a macroscopic view, we use the vertex and human joints as the two ends of the spring. We assume the clothed human in canonical space represents an equilibrium state. So the initial length $L$ of the spring can be viewed as the distance between vertices and force points, which can be considered as joints. Therefore, the displacement $\Delta x_i $  of frame \textit{i} is:
\begin{eqnarray}    \label{eq}
\Delta x_i&=&x_i-L  \nonumber    \\
~&=&(x_i - J_i) - (x_c - J_c) \nonumber    \\
~&=&(x_i - x_c) - (J_i - J_c) .
\end{eqnarray}
Since $x_i$ is the part we need to solve, the formula can be presented as:
\begin{equation}
F_{spring_i}\propto (x_c, \theta_i,T_i,\theta_c,T_c), \label{XX}\\
\end{equation}
where  $x_i$ is  the vertex position at frame \textit{i}, $x_c$ is the vertex position in the canonical space.

Due to the uncontrollable scene, environmental factors such as wind may also affect the motion of the vertices. Therefore, a learnable variable $\varphi$ is added as the effect of the environment on the vertices. As a result, we can get:
\begin{equation}
F_i\propto (x_c,\theta_{i+1},T_{i+1}, \theta_{i-1},T_{i-1}, \theta_i,T_i,\theta_c,T_c,\varphi). \label{XX}\\
\end{equation}

We represent the non-rigid deformation of each frame with a learnable MLP. The point $x'$ deformed by non-rigid deformation field can be represented as:
\begin{equation}
x'=F_i(x_c,\theta_{i+1},T_{i+1}, \theta_{i-1},T_{i-1}, \theta_i,T_i,\theta_c,T_c,\varphi). \label{XX}\\
\end{equation}

\subsubsection{Skinning Transformation Field.} 
Given the \textit{i}-th frame's pose parameter $\theta_i$, we define a canonical-to-current space skinning transformation field. 
Following prior works \cite{jiang2022selfrecon,lin2022learning}, we propagate the SMPL skinning weights in canonical space to get the initial skinning weights of arbitrary topologies. Specifically, we find the 30 nearest vertices on the SMPL mesh in canonical space for each point and average their skinning weights with IDW (inverse distance weight) as the initial weight.

Though using the initial skinning weights method to provide a good beginning, we still need to optimize them for the current subject. Similar to HumanNeRF \cite{weng2022humannerf}, we employ a CNN to learn the weight offsets instead of solving for the entire set of skinning weights: 
\begin{equation}
w= softmax(log(w_{init}) + CNN(x';z)),
\end{equation}
which can achieve faster and more accurate weight optimization.

We use PyMAF \cite{zhang2021pymaf} to extract SMPL parameters from images. PyMAF leverages a feature pyramid and rectifies the predicted parameters explicitly based on the mesh-image alignment status. Although PyMAF improves the alignment between meshes and images on 2D planes, it struggles to tackle the depth ambiguity problem in 3D space. To address this, we introduce an additional network to optimize the human pose. Similar to the skinning weight optimization, we use an MLP to obtain a relative value for pose optimization:
\begin{equation}
\Delta \Omega= MLP_{\theta }(\Omega),
\end{equation}
where $\Omega  = (\omega _0,..., \omega _k)$ are local joint rotations
represented as axis-angle vectors $\omega _i$. We keep the joints \textit{J} fixed and optimize the relative updates of the joint angles, $\Delta \Omega  = (\Delta \omega _0,..., \Delta \omega _k)$. We then apply these updates to $ \Omega $ to obtain the updated rotation vectors:
\begin{equation}
pose(\theta)= (J, \Delta \Omega \otimes  \Omega).
\end{equation}

Finally, the skinning deformation is:
\begin{equation}
x_i= W(x', pose(\theta), w).
\end{equation}

\subsection{Delayed Optimization}
During the optimization of the overall network,  there are several modules and a lot of learnable parameters. As a result, the modules are not decoupled from each other. This leads to suboptimal learning outcomes for each module. Inspired by HumanNeRF \cite{weng2022humannerf}, we deal with this issue by managing the overall optimization process. The optimizations for pose and skinning weights are disabled at the beginning of training, and they are gradually enabled when the non-rigid deformation network acquires a certain level of representation capacity. This approach can effectively alleviate the burden of network learning. Furthermore, pose optimization is only applied in the skinning transformation. Applying pose optimization in non-rigid transformation will increase the complexity of the non-rigid network and produce poor results.

\subsection{Implicit Rendering Network}
To obtain accurate geometry, we use surface rendering instead of volume rendering. While volume rendering can produce good rendering results, it generally yields poor geometry. Following the approach of IDR \cite{yariv2020multiview}, we input surface point, normal, view direction, and global geometric features into an MLP to estimate the colors of surface points. The obtained colors of points fully consider BRDF and global illumination and approximate the surface light field.

We only train the color field in canonical space to reduce memory usage and parameter amount. Followed by SelfRecon \cite{jiang2022selfrecon}, we sample pixels within the ground truth mask and utilize non-rigid ray casting to obtain the corresponding point $x_c$ in canonical space. In the meantime, we compute its normal $n_{x_c}= \bigtriangledown f(x_c)$ by gradient calculation. Given the camera information, we can determine the viewing direction $v$ of each surface point $x_d$ in the current space. By using the Jacobian matrix $J_{x_d}(x_c)$ of the deformation point $x_i=W(F_i(x_c))$, we can transform \textit{v} to the viewing direction $v_c$ of $x_c$ in canonical space.
Finally, we use an MLP to compute the color $L_{x_c}$ of $x_c$, formulated as:
\begin{equation}
L_{x_c}= MLP_{color}(x_c, n_{x_c}, v_{x_c}).
\end{equation}

\subsection{Loss Function}
During the computation of the explicit loss, we regard the canonical mesh $T$ as an optimizable variable and compute its gradient together with the whole network. Then in the consistency loss, we connect explicit variations with the implicit representation. Explicit loss includes mask loss, while implicit losses include color loss, normal loss, and the Eikonal loss.

\textbf{Mask Loss.}
We use the point cloud-based renderer in PyTorch3D 
%\cite{pytorch3d} 
and camera to render out the mask $O'_i$ of the \textit{i}-th frame mesh, and target mask $O_i$ to calculate IoU loss:
\begin{equation}
loss_{IoU} = 1 -  \frac{\left \|O'_i \otimes  O_i\right \| _1}{\left \|O'_i \oplus  O_i -  O'_i \otimes  O_i\right \| _1} ,
\end{equation}
where $\otimes$ and $\oplus$ are the operators that perform element-wise product and sum respectively.

\textbf{Normal loss.}
We use the normal map predicted by PIFuHD \cite{saito2020pifuhd} to refine the geometry. By gradient calculation, we can easily get normal $n_{x_c}$. In addition, we need to convert the corresponding predicted normal $N$ from current space to canonical space, which can be calculated using $J_{x_d}(x_c)^{T}$. Therefore, there is a normal loss:
\begin{equation}
loss_{norm} = \left \| n_{x_c} - unit(J_{x_d}(x_c)^{T} N) \right \| _2,
\end{equation}
where $unit(\cdot )$ means to normalize the vector.

\textbf{Color loss.}
We minimize the color difference between the rendered image $L_i$ and the input frame $I_i$ as:
\begin{equation}
loss_{color} = \left |L_i - I_i \right | .
\end{equation}

\textbf{Eikonal Loss.}
We adopt the regular loss of IGR \cite{gropp2020implicit} to make $f$ to be signed distance function: 
\begin{equation}
loss_{eik} = (\left \| \bigtriangledown f(x_c) \right \| _2 -1)^{2} .
\end{equation}

Finally, the implicit loss can be represented as:

\begin{equation}
\begin{split}
Loss_{Implicit} =loss_{color}+loss_{eik}+\lambda loss_{norm}, 
\end{split}
\end{equation}
where $\lambda=0.1 $.

\textbf{Consistency Loss.}
After explicit iteration, the canonical mesh \textit{T }is updated to $\hat{T}$. To maintain consistent between the implicit SDF $f$ and the updated explicit mesh $\hat{T}$ during implicit iteration, we employ a consistency loss from SelfRecon \cite{jiang2022selfrecon}:

\begin{figure*}[!h]
\centering
\includegraphics[width=1.0\textwidth]{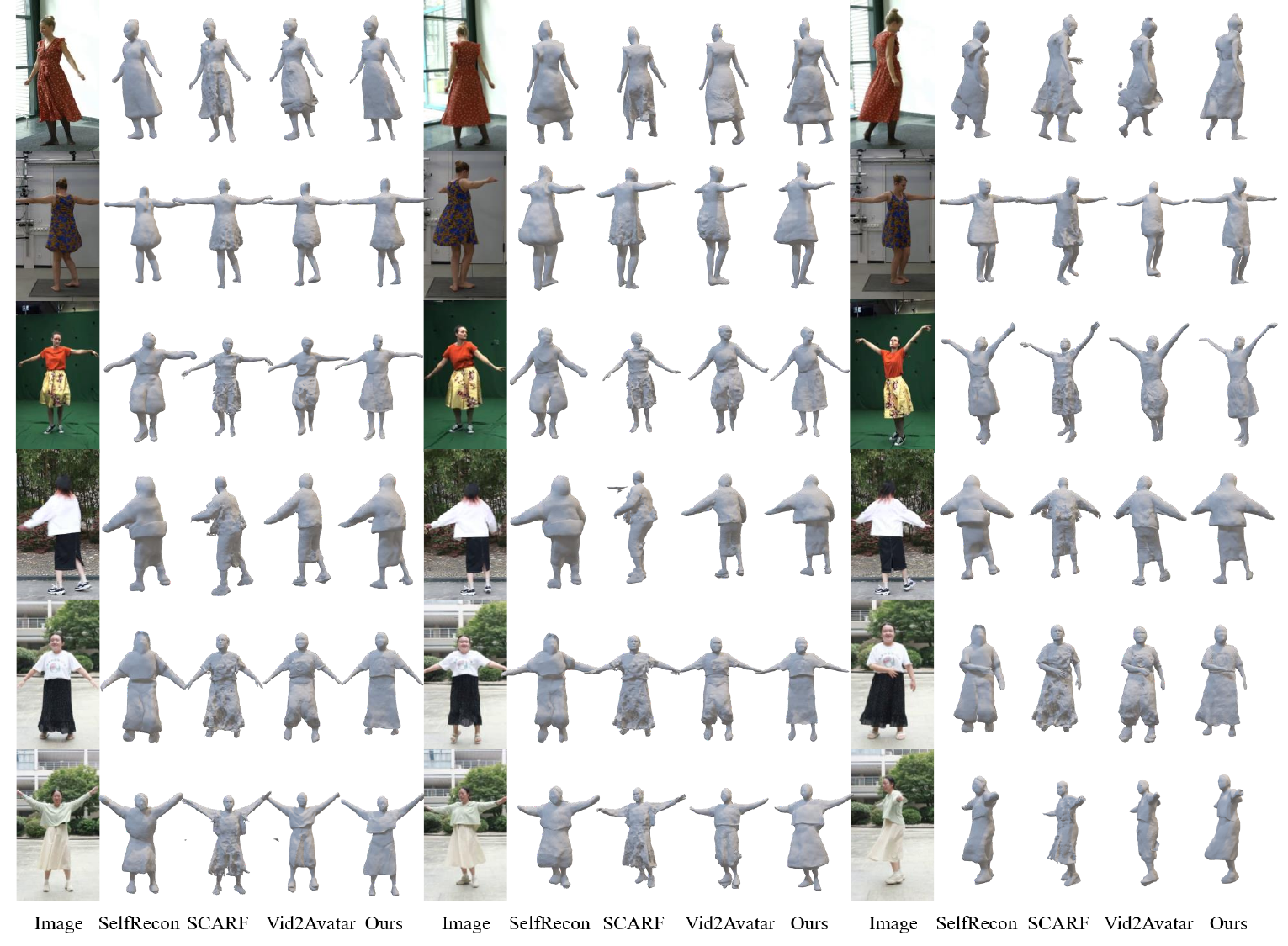} % Reduce the figure size so that it is slightly narrower than the column.
\caption{Geometric qualitative comparison. From top to bottom: ``Antonia'' and ``Magdalena'' from DeepCap Dataset \cite{habermann2020deepcap}, ``FranziRed" from DynaCap dataset \cite{habermann2021real}, and others from self-captured real sequences.}
\label{geo_comparison}
\end{figure*}

\begin{equation}
\begin{split}
Loss_{cons}=\frac{1}{| \hat{T} | } \sum_{\hat{t} \in \hat{T}}^{} \left | f(\hat{t}) \right | 
\end{split}
\end{equation}
where $\hat{t}$ is a vertex coordinate of $\hat{T}$. Intuitively, the loss demands alignment between $\hat{T}$ and the implicit surface. 
%In each optimization step, we start with an explicit iteration to obtain $\hat{T}$ and retain its gradient. Then, we compute implicit and consistency losses to accumulate new gradients. Finally, the network is updated using gradient calculations with Adam.

\begin{table}[t] % ht in default
\centering
\resizebox{\linewidth}{!}{
\begin{tabular}{c|cc|cc}
\hline
\multirow{2}{*}{Subject} & \multicolumn{2}{|c}{Normal MAE$^{*}$  $\downarrow$} & \multicolumn{2}{|c}{Mask IoU $\uparrow$}    \\ 
\cline{2-5}
& SelfRecon & Ours                    & SelfRecon & Ours              \\ 
\hline
Antonia & {11.19} & \textbf{6.14} & 0.887 & \textbf{0.904} \\
Magdalena & {11.40} & \textbf{7.69} & 0.901 & \textbf{0.912}\\
FranziRed& {9.32} & \textbf{5.29}  & \ 0.735 &  \textbf{0.912}\\
LCJ& {11.38} & \textbf{7.63}  & \ 0.884 &  \textbf{0.910}\\
LYZ& {20.71} & \textbf{8.00}  & \ 0.831 &  \textbf{0.919}\\
ZJ& {15.74} & \textbf{9.46}  & \ 0.819 &  \textbf{0.906}\\
\hline
\end{tabular}
}
\caption{Quantitative comparison on geometry. ``$\uparrow$" indicates the higher the better, and ``$\downarrow$" indicates the lower the better. Normal MAE$^{\ast } =$ Normal MAE $\times  10^{3}$ . ``Antonia'' and ``Magdalena'' are from DeepCap Dataset \cite{habermann2020deepcap}, ``FranziRed" is from DynaCap dataset \cite{habermann2021real}, and others are self-captured real sequences.}
\label{table_quan_geo}
\end{table}

\begin{table*}[t] % ht in default
\centering
\resizebox{\linewidth}{!}{
\begin{tabular}{c|cccc|cccc}
\hline
\multirow{2}{*}{Subject} & \multicolumn{4}{c|}{PSNR  $\uparrow$} & \multicolumn{4}{c}{SSIM $\uparrow$}
\\ 
\cline{2-9}
& SelfRecon & HumanNeRF & Vid2Avatar & Ours           
& SelfRecon& HumanNeRF & Vid2Avatar & Ours               \\ 
\hline
Antonia 
& {34.98}& {31.87}& {31.41} & \textbf{37.98} 
& {0.987}& {0.978}& {0.990} & \textbf{0.991} \\
Magdalena 
& {36.07}& {30.14}& {31.79} & \textbf{39.77} 
& {0.987}& {0.974}& {0.990} & \textbf{0.992} \\
FranziRed 
& {31.92}& {32.48}& {32.04} & \textbf{34.09} 
& {0.984}& {0.988}& {0.990} & \textbf{0.991}  \\
LCJ 
& {29.17}& {31.62}& {34.13} & \textbf{34.43} 
& {0.989}& {0.986}& \textbf{0.991} & \textbf{0.991}  \\
LYZ
& {23.04}& {31.69}& {32.49} & \textbf{38.14} 
& {0.966}& {0.981}& \textbf{0.987} & \textbf{0.987} \\
ZJ 
& {31.50}& {32.00}& {35.70} & \textbf{36.94} 
& {0.979}& {0.984}& \textbf{0.989} & {0.986}\\
\hline
\end{tabular}
}
\caption{Quantitative comparison on rendering. ``$\uparrow$" indicates the higher the better, and ``$\downarrow$" indicates the lower the better. ``Antonia'' and ``Magdalena'' are from DeepCap Dataset \cite{habermann2020deepcap}, ``FranziRed" is from DynaCap dataset \cite{habermann2021real}, and others(``LCJ'', ``LJZ'', ``ZJ'') are self-captured real sequences.}
\label{table_quan_render}
\end{table*}

\section{Experiments}

We evaluate our method on the DeepCap dataset \cite{habermann2020deepcap}, DynaCap dataset \cite{habermann2021real} and our own captured data (LCJ, LYZ and ZJ). Our data is captured in the wild with a static CANON EOS 6D MARK \uppercase\expandafter{\romannumeral2} camera. We fix the focal length and estimate the camera intrinsics using COLMAP. For each subject, we use 200-300 images for optimization. The optimization takes 200 epochs (about 48 hours) on a single NVIDIA RTX 3090 GPU. 

\subsection{Quantitative Evaluation}
We utilize normal MAE and mask IoU as evaluation metrics of geometry. The results are presented in Table \ref{table_quan_geo}. We estimate the normal from PIFuHD \cite{saito2020pifuhd} and the mask from RVM \cite{lin2022robust} as ground truth. Table \ref{table_quan_geo} shows that our reconstructed geometry outperforms SelfRecon in terms of both silhouettes and normal.

We report SSIM and PSNR to measure rendering quality. Results in Table \ref{table_quan_render} demonstrate that our method achieves higher accuracy than other methods under most metrics. Our method outperforms SelfRecon \cite{jiang2022selfrecon} and HumanNeRF \cite{weng2022humannerf} in all metrics. Compared to the SOTA method Vid2Avatar \cite{guo2023vid2avatar}, our approach outperforms in loose clothing with a large wiggle amplitude.

\begin{table*} % ht in default
\centering
\resizebox{\linewidth}{!}{
\begin{tabular}{c|ccc|ccc|ccc|ccc}
\hline
\multirow{2}{*}{ } & \multicolumn{3}{c|}{Normal MAE$^{\ast }$  $\downarrow$} & \multicolumn{3}{c}{IoU $\uparrow$}   & \multicolumn{3}{c|}{PSNR  $\uparrow$} & \multicolumn{3}{c}{SSIM $\uparrow$} \\ 
\cline{2-13}
& Antonia & Magdalena & Lab$^{\ast }$ 
& Antonia & Magdalena & Lab$^{\ast }$ 
& Antonia & Magdalena  & Lab$^{\ast }$ 
& Antonia & Magdalena & Lab$^{\ast }$ \\ 
\hline
baseline 
& {11.19}& {11.40} & {18.90}
& {0.887}& {0.901} & {0.858}
& {34.98}& {36.07} & {26.63}
& {0.987}& {0.987}& {0.978}\\
baseline+initialization 
& {6.44}& {8.67} & {10.65}  
& {0.900}& {0.905} & {0.904}
& {37.42}& {37.94} & {32.03}
& {0.985}& {0.983} & {0.982}\\
baseline+initialization+dynamic non-rigid 
& \textbf{6.03}& {7.93} & {10.22}
& {0.903}& {0.910} & {0.916} 
& {37.68}& {38.80} & {33.31}
 & {0.985}& {0.984} & {0.983}\\
 Ours (w/o dynamic non-rigid) 
& {6.65}& {8.83} & {10.90}  
& {0.895}& {0.904} & {0.903} 
& {37.01}& {37.96} & {32.03}
& {0.985}& {0.983}& {0.982}\\
Ours (full model) 
& {6.14}& \textbf{7.69} & \textbf{10.22} 
& \textbf{0.904}& \textbf{0.912} & \textbf{0.921}  
 &\textbf {37.98}& \textbf{39.77} & \textbf{33.81}
 & \textbf{0.990}& \textbf{0.992} & \textbf{0.985}\\

\hline
\end{tabular}
}
\caption{Ablation study of geometry and rendering.``$\uparrow$" indicates the higher the better, and ``$\downarrow$" indicates the lower the better. Normal MAE$^{\ast } =$ Normal MAE $\times  10^{3}$ . We compute averages over 3 sequences of Lab Dataset.  Lab Dataset contains self-captured video clips. ``Ours (w/o dynamic non-rigid)" means we use another non-rigid deformation field with only frame index. Our full model contains initialization, dynamic non-rigid and optimized skinning deformation fields, as well as a pose decoder.}
\label{table_ablation_geo}
\end{table*}

\subsection{Qualitative Evaluation}
We also conduct qualitative comparisons with SelfRecon \cite{jiang2022selfrecon}, SCARF \cite{feng2022capturing} and Vid2Avatar \cite{guo2023vid2avatar} on the DeepCap Dataset \cite{habermann2020deepcap} and our collected real sequences in Figure \ref{geo_comparison}. More results are in the supplementary \footnote{https://github.com/surheaven/DLCA-Recon}.

SelfRecon tends to produce physically incorrect body reconstructions. Due to the inherent limitations of NeRF, SCARF tend to reconstruct noisy geometry. SCARF proposes a hybrid model combining a mesh-based body with a NeRF-based clothing. Although it obtains relatively clean bodies, it reconstructs clothing with a lot of noise. This may be due to its inability to capture clothing dynamics in free motion. Following HumanNeRF and NeuMan \cite{jiang2022neuman}, Vid2Avatar reconstructs an avatar via self-supervised scene decomposition. Though Vid2Avatar performs well on garments that are topologically similar to the body, it still fails to reconstruct loose clothing. It struggles to reconstruct loose-fitting clothing due to their fast dynamics.  
In contrast, our method generates complete and accurate results regardless of whether the clothing is tight or loose-fitting. %More results see in supplementary.

%\begin{figure}[t]
%\centering
%\includegraphics[width=1.0\columnwidth]%{AnonymousSubmission/LaTeX/images/colorcomparison1_clip.pdf} 
%\caption{Rendering qualitative comparison.}
%\label{color_comparison}
%\end{figure}

\begin{figure}[htb]
\centering
\includegraphics[width=1.0\columnwidth]{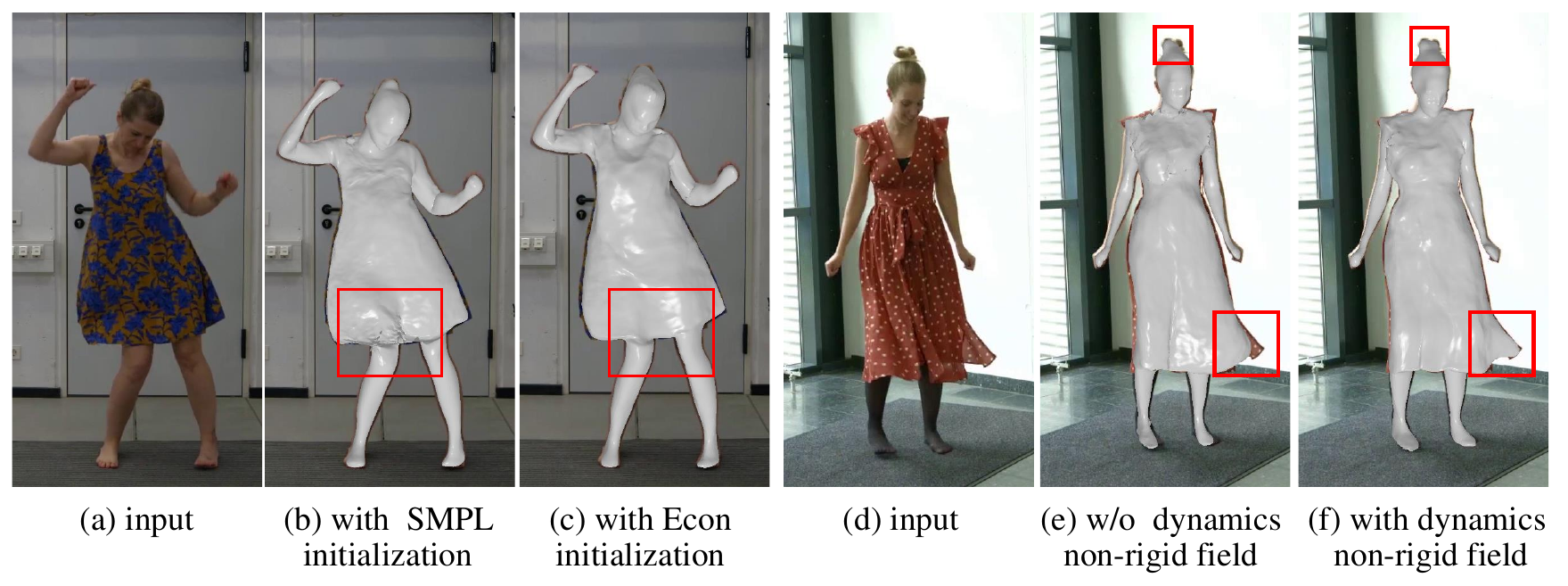} 
\caption{SMPL initialization cannot expand to clothes that are not similar to the body's topology (b). Econ initialization gives the correct topology for the clothed human (c). The dynamics non-rigid deformation improves clothing alignment and shape (d-f).
%In this case, SMPL initialization tends to generate the incorrect  geometry.
}
\label{ablations}
\end{figure}

\subsection{Ablation Study}

\subsubsection{Effect of Initialization.}
To reconstruct human wearing various types of clothing, we select a frame from the video and employ ECON \cite{xiu2023econ} to derive an initial geometric. Table \ref{table_ablation_geo} shows the ablation experiments using ECON and SMPL initialization. As demonstrated in the results, when using SMPL initialization, even in conjunction with the SDF-based implicit method, it is prone to get stuck in local optima and cannot generate loose-fitting clothing. Figure \ref{ablations} shows that the initialization of ECON lays the foundation for avatar reconstruction. Effective initialization gives the correct topology and facilitates network learning.

\subsubsection{Effect of Dynamic Non-Rigid Deformation Field.}
Table \ref{table_ablation_geo} shows that the dynamic non-rigid field can better capture clothing movement, especially in the case of loose clothing. It proves the validity of force formulation in a non-rigid deformation field. Figure \ref{ablations} visualizes the importance of including dynamics non-rigid motion. To verify the force formulation in the dynamic non-rigid field, we conduct an ablation study between dynamic non-rigid MLP and a non-rigid MLP with only frame index in Table \ref{table_ablation_geo}.

\subsubsection{Effect of Optimized Skinning Deformation Field.}
In Table \ref{table_ablation_geo}, we find that using weights fine-tuning can get more correct geometry and improve the metrics to some extent. In a word, dynamics non-rigid alone is enough for significant improvement. Adding LBS fine-tuning provides further gains. We also conduct ablation experiments on the pose decoder and delayed optimization (see supplementary). 

\subsubsection{Effect of SMPL Pose Estimation Method.}
We replace PyMAF \cite{zhang2021pymaf} with a slightly non-robust method called SPIN \cite{kolotouros2019learning}, which is compared in PyMAF work. Table \ref{table_spin} shows that our method can get stable accurate results with SPIN.
As we employ a pose decoder to refine poses and apply a skeleton smoothness loss to maintain low-frequency joint motion trajectories, these measures decrease our sensitivity to SMPL poses.

\begin{table}[!h] % ht in default
\centering
\resizebox{\linewidth}{!}{
\begin{tabular}{c|c|c|c|c}
\hline
& {Normal MAE$^{\ast }$   $\downarrow$ } 
& {IoU $\uparrow$} 
& {PSNR  $\uparrow$} 
& {SSIM $\uparrow$}     \\ 
\cline{2-5}
 
\hline
FranziRed(PyMAF) & {5.29}  & {0.912} & {34.09}  & {0.991} \\
FranziRed(SPIN) & {5.31}  & {0.912} & {34.04}  & {0.991} \\

\hline
\end{tabular}
}

\caption{Ablation study of SMPL pose estimation method. }
\label{table_spin}
\end{table}

\section{Conclusion and Discussion}
In this paper, we propose a method named DLCA-Recon to reconstruct 3D avatars from monocular in-the-wild videos. By employing force formulation in non-rigid deformations and optimizing skinning weights through initialization, we can effectively capture the free motion of bodies and clothes. Managing the overall network optimization process helps mitigate the coupling between modules to a certain extent. Without the requirement of scans as supervision, DLCA-Recon can reconstruct high-fidelity humans dressed in a variety of clothing styles from monocular videos in the wild.

DLCA-Recon still has several limitations. First, due to employing surface rendering loss, our method is limited by the accuracy of the ground truth mask. Second, current approaches rely on predicted normal maps to improve geometric details. Lastly, the geometry obtained from 2D supervision is still inferior to 3D supervision. %Lastly, despite employing methods to manage optimization, numerous modules and parameters prevents complete decoupling of each module, resulting in mutual influence among them. 
We will address these issues in the future.

\section{Acknowledgments}
This work is partially supported by the National Natural Science Foundation of China
(No. 61972298 and No. 62372336), CAAI-Huawei MindSpore Open Fund and Wuhan University-Huawei GeoInformatices Innovation Lab.

\appendix
\section{Appendices}

\section{A. Implementation details}
\subsection{A.1 Corresponding Points between Canonical Space and Current Space.}
We transform the point $x_c$ from canonical space to current frame space ($i$-th frame) and get deformed points $x_i$.
We need to find the corresponding points between two spaces to utilize color loss. Following SelfRecon \cite{jiang2022selfrecon}, we use differentiable non-rigid ray-casting. Specifically, given a ray r with camera position c and ray direction v, we can get the first intersection $p_i$ on the deformed mesh. Moreover, with the intersected triangle on the deformed mesh, we can find $p_c$'s corresponding point $p_c$ on the canonical mesh by consistent barycentric weights. With $p_c$ as the good initialization of $x_c$, we deform point $x_i = D_i(x_c)=W(F_i(x_c)) $ and $x_i$ is the intersection point of the ray r and the current space SDF. The SDF in canonical space is $f$, and  $p_c$ is on the surface of canonical mesh, so we need to drive $f(p_c)$ to 0.
Specifically, we solve p by:

\begin{eqnarray}    \label{eq}
x_c&=&\mathop{\min}   
\left | f(p_c) \right |   +  
\frac{\left \|(p_i- c) \times v \right \| _2}
{ \left \|p_i- c \right \|_2 } \nonumber    \\
~&=&\mathop{\arg\min}_{p_c}   
\left | f(p_c) \right |   +  
\frac{\left \|(D_i(p_c)- c) \times v \right \| _2}
{ \left \|(D_i(p_c)- c \right \|_2 }  .
\end{eqnarray}

\begin{figure}[!h]
\centering
\includegraphics[width=1.0\columnwidth]{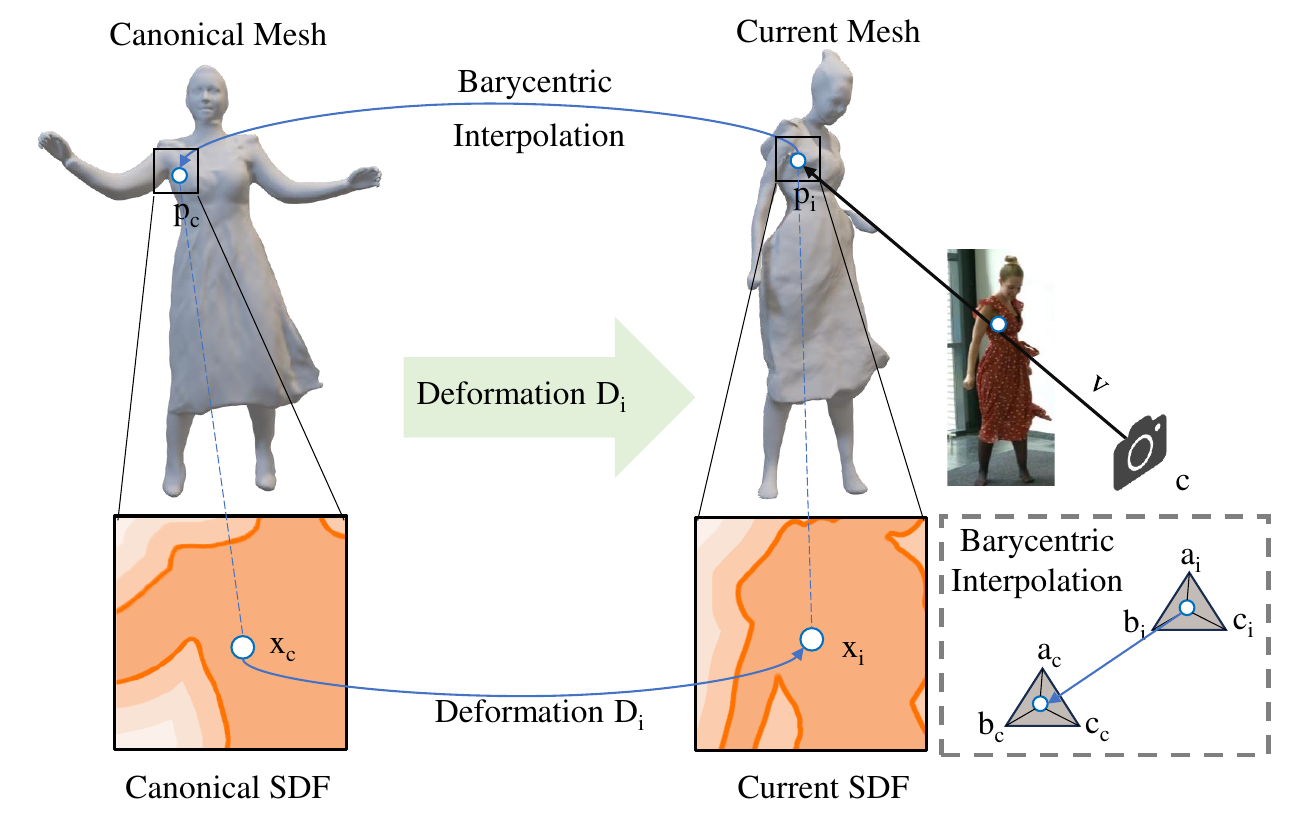} 
\caption{We render the mesh in the current space and find visible faces. We get the point $p_{i}$ on one visible face and barycentric weights. Then we use the barycentric interpolation to initialize the corresponding point $p_c$ in canonical space.
$\bigtriangleup a_{i}b_{i}c_{i}$ is one visible face in the current space. Moreover, the $\bigtriangleup  a_{c}b_{c}c_{c}$ is the corresponding face of $\bigtriangleup a_{i}b_{i}c_{i}$ in canonical space. }
\label{corresponding_points}
\end{figure}

\subsection{A.2 More Details of Initialization.}
\begin{figure}[!h]
\centering
\includegraphics[width=1.0\columnwidth]{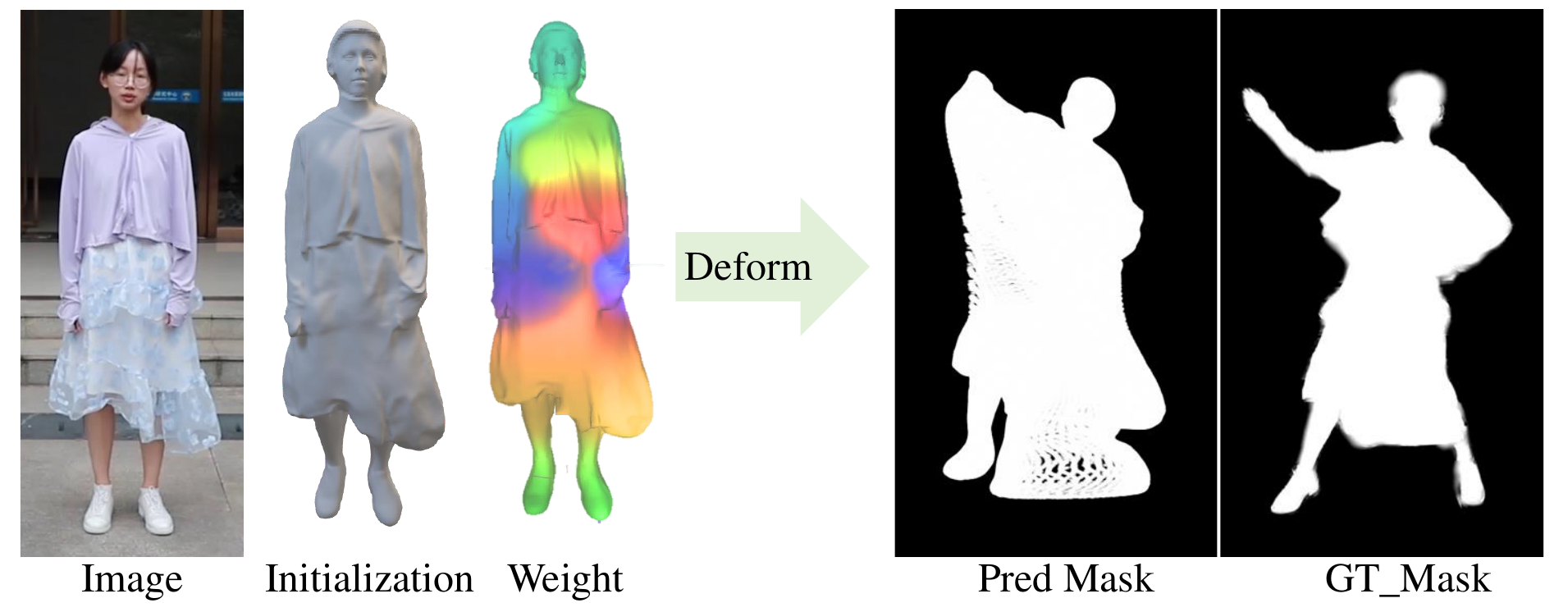} 
\caption{We select a frame for initialization. The initialization with limbs not separated leads to wrong initial weights and incorrect human body deformation.  }
\label{bad_ini}
\end{figure}

We can initialize from any frame of the video. Following NeuMan \cite{jiang2022neuman}, we prefer the frame where the limbs are separated for initialization. It avoids collision when warping from canonical space to current space. In Figure \ref{bad_ini}, the arms and the body are merged in the initialization. It leads to incorrect initial weights. We can see that the weight of the hand (purple) is incorrectly spread to the skirt. Additionally, forward deformation cannot be effectively performed when limbs are not separated, resulting in geometric distortions.

\subsection{A.3 Network Architecture.}
\begin{figure}[!h]
\centering
\includegraphics[width=1.0\columnwidth]{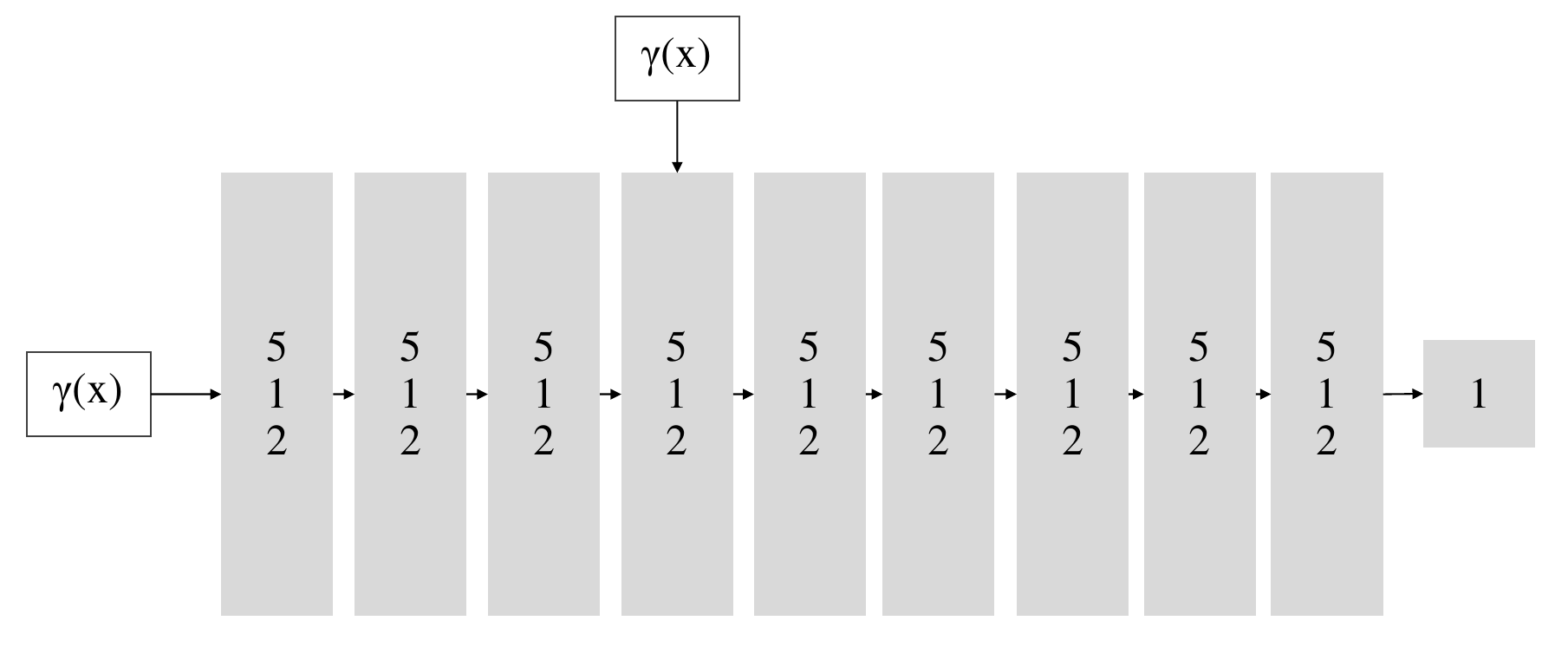} 
\caption{Canonical human shape SDF visualization.  }
\label{SDF_visualization}
\end{figure}

\begin{figure}[!h]
\centering
\includegraphics[width=1.0\columnwidth]{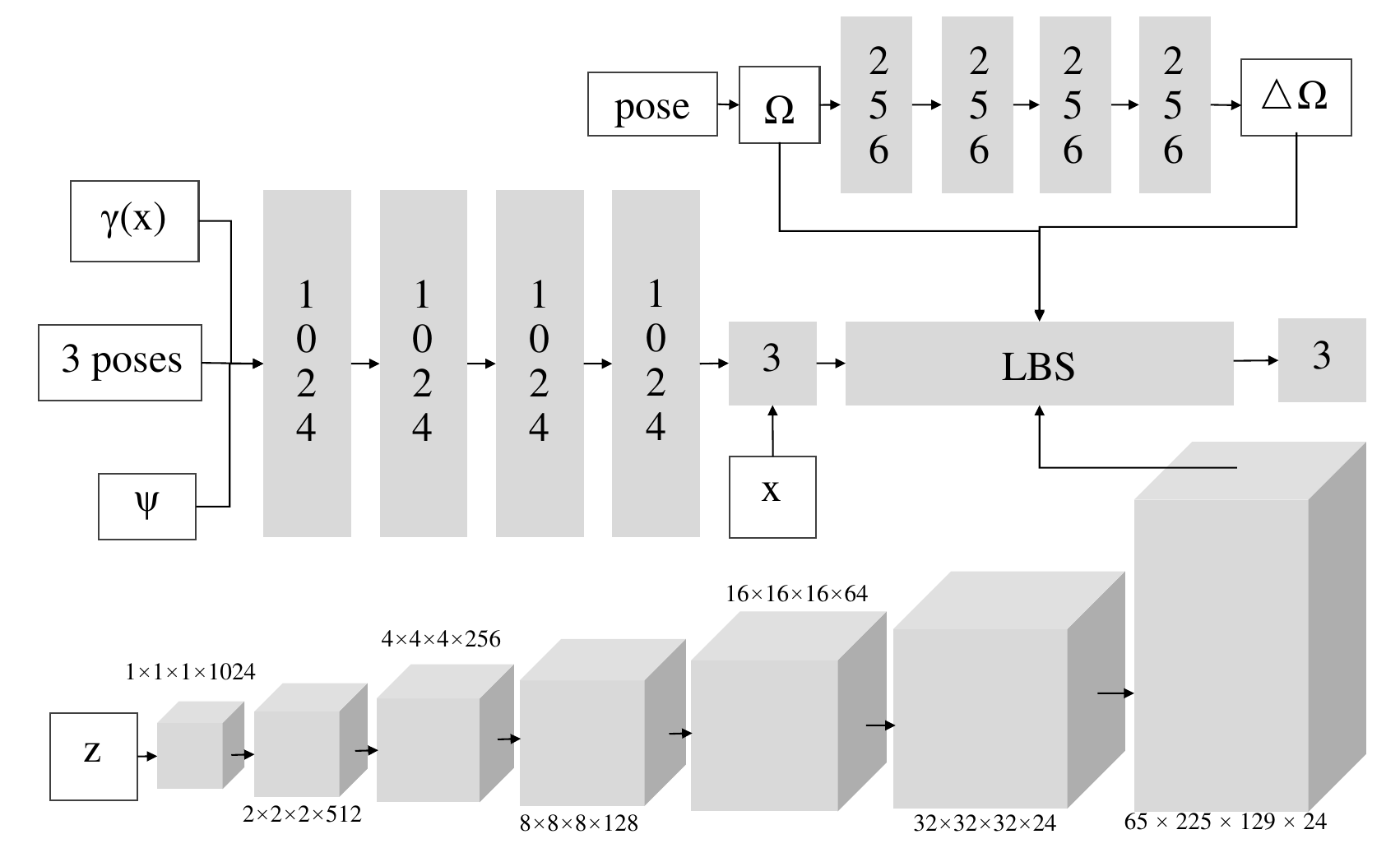} 
\caption{The visualization of dynamic deformation field DDF. DDF contains the dynamic non-rigid deformation field (middle), the pose decoder (top), and the weights finetune network (down). }
\label{DDF_visualization}
\end{figure}

As Figure \ref{SDF_visualization} shows, the canonical human shape SDF network $f$ includes 9 fully connected layers, each followed by a softplus activation layer. In Figure \ref{DDF_visualization}, the dynamic deformation field DDF consists of a dynamic non-rigid field and an optimized skinning transformation field. The non-rigid field comprises 5 linear layers, each of which is terminated by a RELU activation. We apply LBS skinning transformation from SMPL \cite{loper2015smpl} and optimize the skinning weights in the skinning transformation field. The weights optimization network starts with a fully-connected layer that transforms the latent code to a 1 × 1 × 1 × 1024 grid. Subsequently, it is combined with 6 transposed convolutions, progressively increasing the volume size while reducing channel count, and finally produces a volume of size 65 × 225 × 129 × 24. LeakyReLU is applied after MLP and transposed convolution layers. Pose finetune network input joint angles $\Omega$ into a 5-layer MLP with a width of 256 and output $\bigtriangleup \Omega $. The implicit rendering network is a 9-layer MLP finishing with a softplus activation layer.

\section{B. More Results}
\subsection{B.1 Comparison with ECON.}
\begin{figure}[!h]
\centering
\includegraphics[width=1.0\columnwidth]{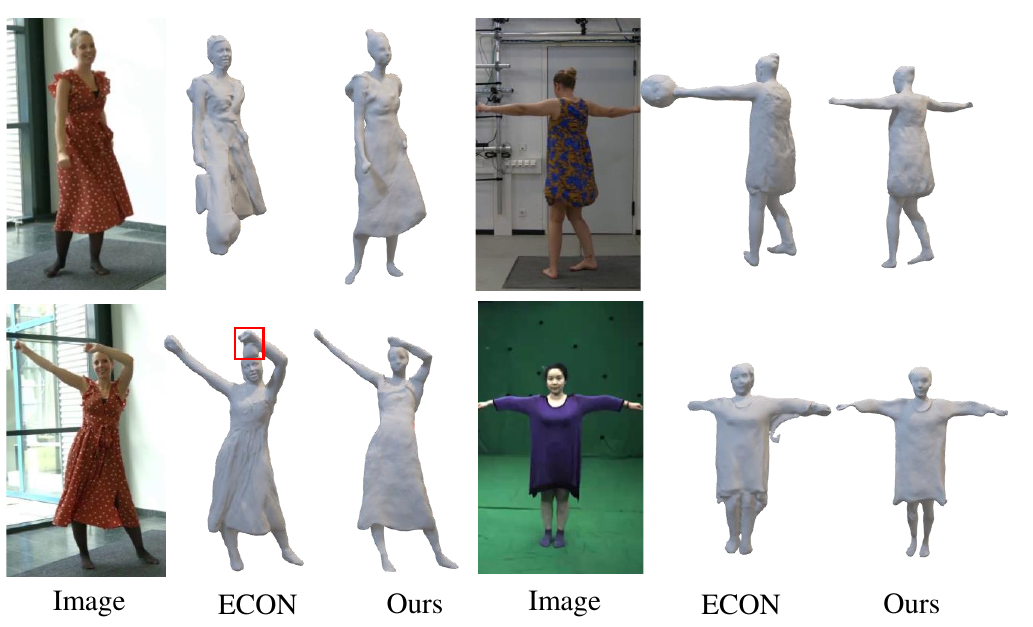} 
\caption{Geometric qualitative comparison of ECON.
The first row demonstrates that ECON leads to reconstruction errors due to inaccurate SMPL pose estimation and mask splitting. The second row shows that, even with a similar and correct SMPL pose, ECON may still yield bad results.}
\label{econ_comparison}
\end{figure}

In Figure \ref{econ_comparison}, we conducted qualitative comparisons of Econ \cite{xiu2023econ}.  As we can see, ECON is limited by the normal and pose predictions, which makes it prone to obtaining incomplete and erroneous results. We also adopt an indirect experimental approach by comparing the mesh reconstruction when ECON and ours obtain similar and correct SMPL to ensure fairer comparisons.
In contrast, our method can generate complete and correct clothed avatars.

\subsection{B.2 Additional Qualitative Studies.}

\begin{figure*}[!h]
\centering
\includegraphics[width=1.0\textwidth]{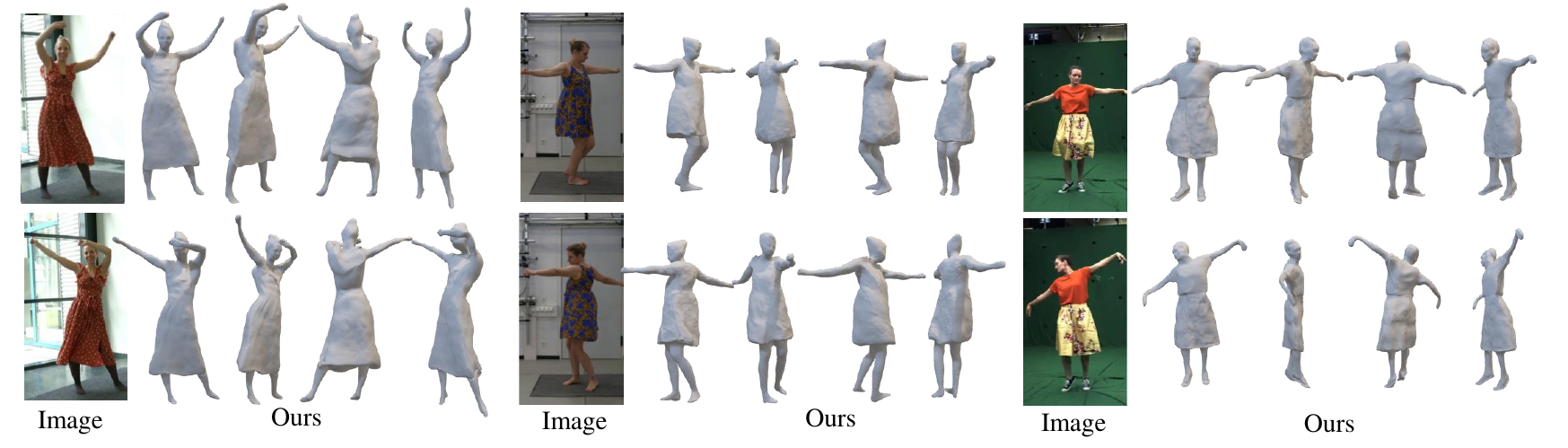} 
\caption{We show four perspectives of the clothed human. From left to right: ``Antonia'' and ``Magdalena'' from DeepCap Dataset \cite{habermann2020deepcap}, ``FranziRed" from DynaCap dataset \cite{habermann2021real}. }
\label{four_views}
\end{figure*}

We also provide four views of estimated meshes in Figure \ref{four_views}. The unusual waist and arms result from pose errors caused by occlusion or invisibility. Existing pose estimation methods and 2D supervised avatar reconstruction methods cannot address this issue. Due to the 2D supervision inner limitation, it may cause some local unusual results. 

\begin{figure*}[!h]
\centering
\includegraphics[width=1.0\textwidth]{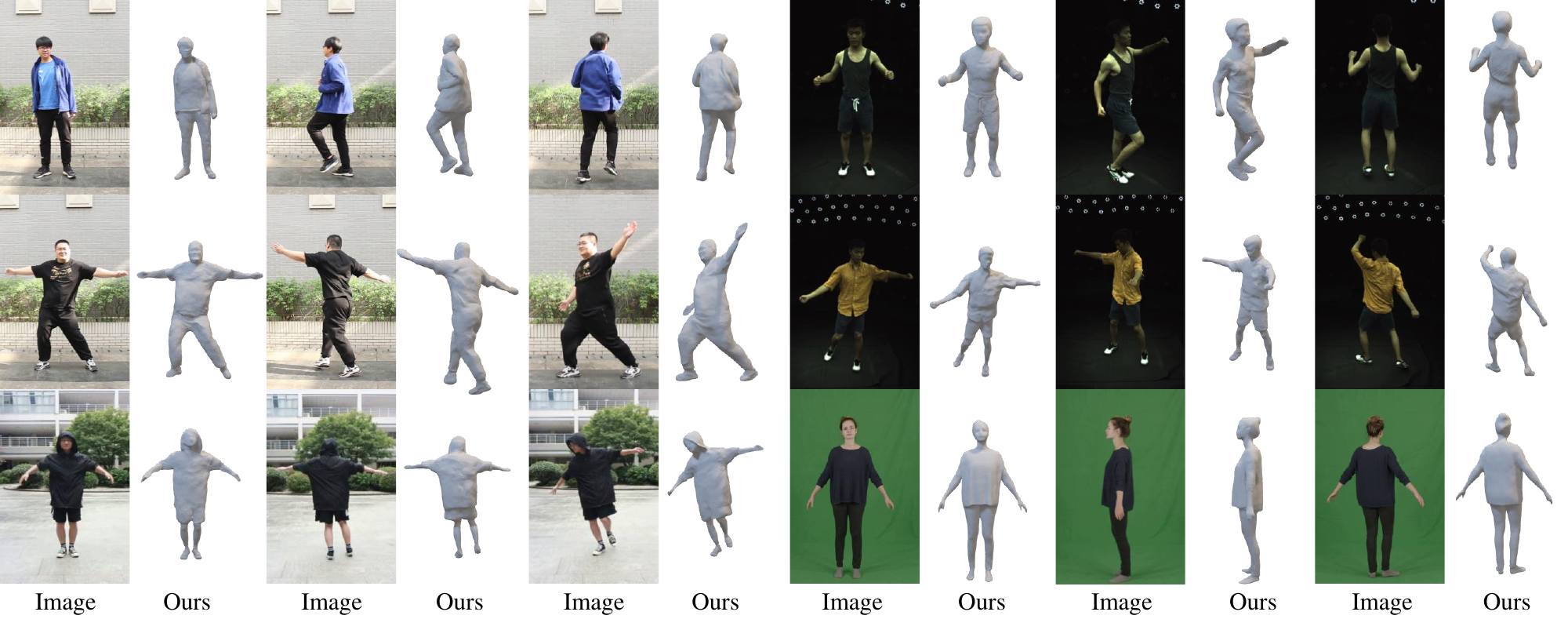} % Reduce the figure size so that it is slightly narrower than the column.
\caption{Additional qualitative geometric results. We have experiments on various types of clothing and DLCA-Recon consistently generates high-quality results. Each group shows images of the video and corresponding reconstructions. The results on the left are self-captured videos, the top two rows on the right are subject 377 and 393 from ZJU-MoCap dataset \cite{peng2021neural}, and the bottom row on the right are ``female-3-casual'' from PeopleSnapshot dataset \cite{alldieck2018video}.}
\label{geo_results_more}
\end{figure*}

We provide additional qualitative geometric results in Figure \ref{geo_results_more}. We conducte experiments on types of clothing, and Figure \ref{geo_results_more} shows that DLCA-Recon can generate favorable results across a variety of clothing types.

\subsection{B.3 Additional Quantitative Studies.}
We make comparisons on public datasets. We compare with SCARF \cite{feng2022capturing} and HumanNeRF \cite{weng2022humannerf} respectively. Table \ref{table_quan_render_scarf} and Table \ref{table_quan_render_humannerf} show that we achieve better performance quantitatively.

\begin{table}[!h] % ht in default
\centering
\resizebox{\linewidth}{!}{
\begin{tabular}{c|cc|cc}
\hline
\multirow{2}{*}{Subject} 
& \multicolumn{2}{|c}{PSNR  $\uparrow$} 
& \multicolumn{2}{|c}{SSIM $\uparrow$}  \\ 
\cline{2-5}
& SCARF & Ours                    
& SCARF & Ours                \\ 
\hline
male-3-casual & {30.59} & \textbf{37.30}
& {0.977} & \textbf{0.984}  \\
male-4-casual & {31.79} & \textbf{38.90} 
& {0.970} & \textbf{0.980} \\
female-3-casual & {30.14} & \textbf{42.63} 
& {0.977} & \textbf{0.987} \\
female-4-casual & {29.96} & \textbf{38.53}
& {0.972} & \textbf{0.976}  \\
\hline
\end{tabular}
}
\caption{Quantitative comparison of SCARF in PeopleSnapshot dataset \cite{alldieck2018video}. We directly copy the metrics from SCARF. ``$\uparrow$" indicates the higher the better.}
\label{table_quan_render_scarf}
\end{table}

\begin{table}[!h] % ht in default
\centering
\resizebox{\linewidth}{!}{
\begin{tabular}{c|cc|cc}
\hline
\multirow{2}{*}{Subject} 
& \multicolumn{2}{|c}{PSNR  $\uparrow$} 
& \multicolumn{2}{|c}{SSIM $\uparrow$}  \\ 
\cline{2-5}
& HumanNeRF & Ours                    
& HumanNeRF & Ours                \\ 
\hline
ZJU-MoCap dataset* 
& {30.24} & \textbf{31.10}
& {0.974} & \textbf{0.978}  \\

\hline
\end{tabular}
}
\caption{Quantitative comparison of HumanNeRF on ZJU-MoCap dataset \cite{peng2021neural}. We directly copy the metrics from HumanNeRF. We only compute ``377", ``386", ``387", ``392", ``393" and ``394" from ZJU-MoCap dataset.
``$\uparrow$" indicates the higher the better.}
\label{table_quan_render_humannerf}
\end{table}

We present the comparison result with SOTA methods on both parameters’ count and inference time in Table \ref{table_compute_complex}. When we get the best reconstruction results, we use a CNN to optimize the deformation field, which increases the parameters’ count. However, it does not increase the inferring time a lot.

\begin{table}[!h] % ht in default
\centering
\resizebox{\linewidth}{!}{
\begin{tabular}{c|c|c}
\hline
Method
& {Params(M)} & {Infer Time(s)}    \\ 
\cline{2-3}
 
\hline
SCARF(NeRF-based Method) & {21.23}  & {16.88} \\
Vid2Avatar & {1.42}  & {992.98} \\
SelfRecon & {3.79}  & {66.51} \\
Ours & {73.45}  & {80.35} \\
\hline
\end{tabular}
}
\caption{Computation complexity comparison on methods.}
\label{table_compute_complex}
\end{table}

\subsection{B.4 Additional Ablation Studies.}

%\subsubsection{Effect of Force Formulation.}

%We introduced the force formulation in non-rigid deformation field for better clothing deformation fitting. To verify its effectiveness, we compared with another non-rigid deformation field with only frame index. Table \ref{table_ablation_non_rigid} shows our method can utilize inputs of force formulation and achieve better results both in geometry and rendering.

\subsubsection{Effect of Pose Decoder.}
\begin{figure}[!h]
\centering
\includegraphics[width=0.9\columnwidth]{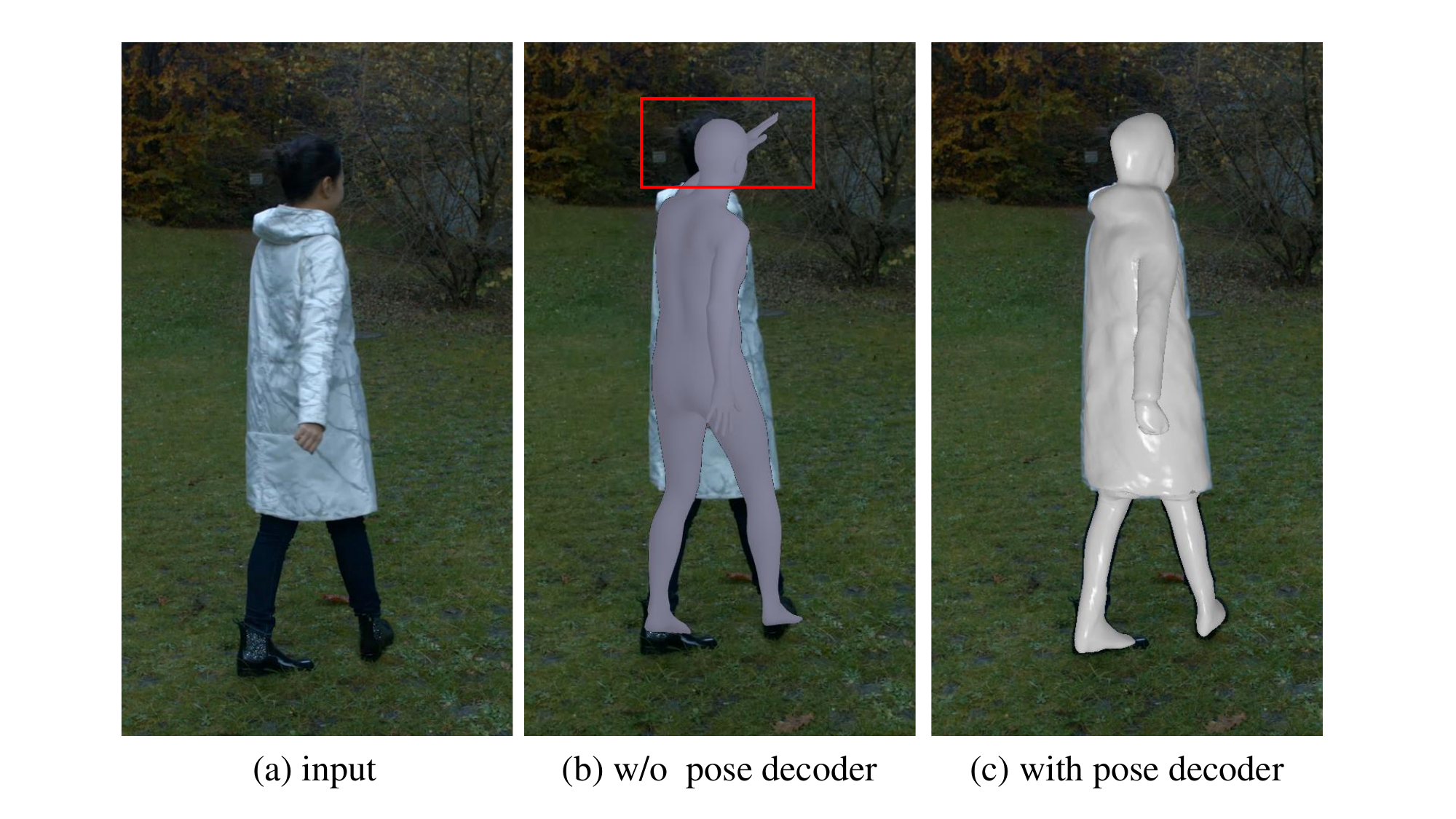} % Reduce the figure size so that it is slightly narrower than the column. Don't use precise values for figure width.This setup will avoid overfull boxes.
\caption{Pose decoder refines the body pose during optimization. It corrects  the left arm from (b) to (c).}
\label{ablation_posedecoder}
\end{figure}

As shown in Figure \ref{ablation_posedecoder}, pose decoder can alleviate the impact of errors in the estimated pose from PyMAF \cite{zhang2021pymaf}. By using accurate poses, we can achieve more precise reconstruction results.

\begin{figure}[!h]
\centering
\includegraphics[width=1.0\columnwidth]{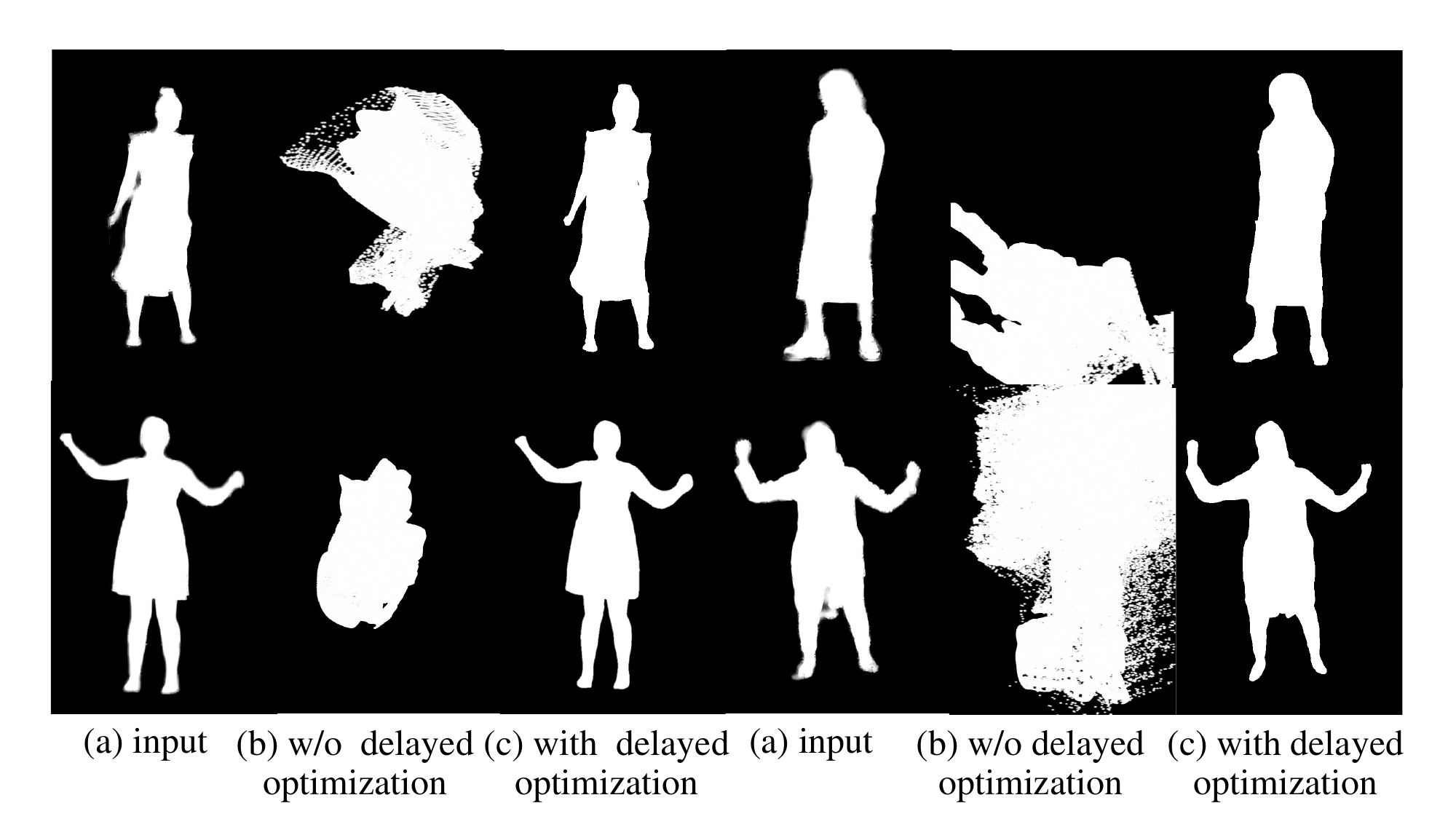} % Reduce the figure size so that it is slightly narrower than the column. Don't use precise values for figure width.This setup will avoid overfull boxes.
\caption{Without delayed optimization, Geometries in (b) become distorted during the training process.}
\label{ablation_delayedop}
\end{figure}

\subsubsection{Effect of Delayed Optimization.}
Figure \ref{ablation_delayedop} demonstrates that without delayed optimization, the reconstruction method fails due to the excessive number of learnable parameters.

%\bigskip
%\noindent Thank you for reading these instructions carefully. We look forward to receiving your electronic files!

\bibliography{aaai24}

\end{document}